\documentclass[conference]{IEEEtran}
\IEEEoverridecommandlockouts
\usepackage{cite}
\usepackage{amsmath,amssymb,amsfonts}
\usepackage{stfloats}
\usepackage{algpseudocode,algorithm,algorithmicx}
\algrenewcommand\algorithmicrequire{\textbf{Inputs:}}  
\algrenewcommand\algorithmicensure{\textbf{Outputs:}}
\renewcommand{\algorithmiccomment}[1]{\bgroup\hfill//~#1\egroup}
\usepackage{graphicx}
\usepackage{float}
\usepackage{textcomp}
\usepackage{tablefootnote}
\usepackage{hyperref}
\usepackage[perpage]{footmisc}
\usepackage{xcolor}
\usepackage{booktabs}
\def\BibTeX{{\rm B\kern-.05em{\sc i\kern-.025em b}\kern-.08em
    T\kern-.1667em\lower.7ex\hbox{E}\kern-.125emX}}
\begin{document}

\algdef{SE}[SUBALG]{Indent}{EndIndent}{}{\algorithmicend\ }%
\algtext*{Indent}
\algtext*{EndIndent}
\newcommand{\Yue}[1]{{\color{blue} [Yue comments ``#1'']}}

\title{COPOD: Copula-Based Outlier Detection}

\author{

\IEEEauthorblockN{Zheng Li}
\IEEEauthorblockA{
\textit{Northeastern Univ. \&}\\
 Arima Inc., Canada \\
winston@arimadata.com}
\and
\IEEEauthorblockN{Yue Zhao}
\IEEEauthorblockA{
\textit{Carnegie Mellon Univ.}\\
Pittsburgh, USA \\
zhaoy@cmu.edu}
\and
\IEEEauthorblockN{Nicola Botta}
\IEEEauthorblockA{
\textit{PIK Potsdam}\\
Germany \\
botta@pik-potsdam.de}
\and
\IEEEauthorblockN{Cezar Ionescu}
\IEEEauthorblockA{
\textit{THD}\\
Germany \\
cezar.ionescu@th-deg.de}
\and
\IEEEauthorblockN{Xiyang Hu}
\IEEEauthorblockA{
\textit{Carnegie Mellon Univ.}\\
Pittsburgh, USA \\
xiyanghu@cmu.edu}
}

\maketitle

\begin{abstract}
Outlier detection refers to the identification of rare items that are deviant from the general data distribution. Existing approaches suffer from high computational complexity, low predictive capability, and limited interpretability. As a remedy, we present a novel outlier detection algorithm called COPOD, which is inspired by copulas for modeling multivariate data distribution. COPOD first constructs an empirical copula, and then uses it to predict tail probabilities of each given data point to determine its level of ``extremeness". Intuitively, we think of this as calculating an anomalous p-value. This makes COPOD both parameter-free, highly interpretable, and computationally efficient. 
In this work, we make three key contributions,
1) propose a novel, parameter-free outlier detection algorithm with both great performance and interpretability, 2) perform extensive experiments on 30 benchmark datasets to show that COPOD outperforms in most cases and is also one of the fastest algorithms, and 3) release an easy-to-use Python implementation for reproducibility.
\end{abstract}


\section{Introduction}
Outliers, also known as anomalies, are data points that have different characteristics from normal observations. Because the existence of outliers can markedly impact the results of statistical analyses, it thus becomes pivotal to identify and remove anomalous data objects from the general data distribution in order to preserve the integrity of data science models. 
Outlier detection algorithms have many practical applications \cite{zhao2019pyod}, such as credit card fraud detection, network intrusion detection, and synthetic data generation \cite{wan2020sync}. These applications demand
outlier detection algorithms with high detection performance, fast execution, and great interpretability.

Numerous algorithms have been proposed over the last decades, which includes covariance based models like Outlier Detection with Minimum Covariance Determinant (MCD) \cite{mcd}, proximity based models like k-Nearest Neighbors Detector (kNN) \cite{Ramaswamy2000efficient}, linear models like One-class SVM detector \cite{scholkopf2001estimating}, ensemble base models like Locally Selective Combination of Parallel Outlier Ensembles (LSCP) \cite{zhao2019lscp}, and neural network based models like Single-Objective Generative Adversarial Active Learning (SO-GAAL) and Multi-Objective Generative Adversarial Active Learning (MO-GAAL) \cite{Liu2019generative}. Existing approaches faces a few limitations. First, their performances suffer from the curse of dimensionality, and are constrained to low dimensional data. Secondly, most methods require hyperparameter selection and tuning (e.g., number of clusters for clustering based models, layer architecture for neural network based models \cite{li2020autood}, and choices for individual classifiers for ensemble models). The choice of hyperparameters is sometimes subjective, leading to different outcomes.

The proposed method, called \textbf{COPOD} (\textbf{COP}ula-based \textbf{O}utlier \textbf{D}etector), has the follow key advantages:

\begin{itemize}
    \item \textbf{COPOD is deterministic without hyperparameters}. COPOD is based on empirical cumulative distribution functions (ECDFs), and involves no learning or stochastic training. This avoids challenges in hyperparameters selection and potential biases. 
    \item \textbf{COPOD is one of the top performing outlier detection algorithms}. Among the 10 popular detectors in comparison, COPOD ranks first, and scores 1.5\% more in ROC-AUC and 2.7\% more in Average Precision than the second best performing detector. 
    \item \textbf{COPOD is highly interpretable via quantifying the abnormality contribution of each dimension through the Dimensional Outlie Graph}. This provides guidance to practitioners on the causes of abnormality---they can focus on certain subspaces to better improve data quality.
    \item \textbf{COPOD is an efficient model, and an ideal choice for high dimensional datasets}. Of the 30 datasets tested, COPOD averages third in execution time. Furthermore, COPOD scales well for large datasets, as low computation overhead is required. COPOD can easily handle datasets with 10,000 features and 1,000,000 observations on a standard personal laptop.
\end{itemize}


\section{Proposed Algorithm}
\label{sec:proposed_algorithm}
This section is divided into two parts. First, we give a theoretical background of copula models, which is the foundation and main motivation of COPOD, and in the second part, we describe the algorithmic design of COPOD in detail.

\subsection{Copula Models}
Copulas are functions that enable us to separate marginal distributions from the dependency structure of a given multivariate distribution. Formally, a $d$-variate copula, $C: [0,1]^d \rightarrow [0,1]$, is the cumulative distribution function (CDF) of a random vector $(U_1, U_2,...,U_d)$ with Uniform$(0,1)$ marginals:
\begin{equation}\label{cop_main}
    C_{\textbf{U}}(\textbf{u}) = \mathbb{P}(U_1 \leq u_1,...,U_d \leq u_d)
\end{equation}
where $P(U_j \leq u_j) = u_j$ for $j \in {1,..,d}$ and $u_j \in [0,1]$. We will drop the subscript $\textbf{U}$ when there is no confusion.

It is well known that uniform distributions can be transformed into any desired distributions via inverse sampling
\begin{equation}\label{inverse relationship}
    X_j = {F_j}^{-1}(U_j) \sim F_j
\end{equation}
Sklar \cite{sklar1959fonctions} show that for any random variables $(X_1,\cdots,X_d)$ with joint distribution function $F(x_1,...,x_d)$ and marginal distributions $F_1,...,F_d$, there exists a copula such that
\begin{equation}\label{sklar1}
    F(\textbf{x}) = C(F_1(x_1),...,F_d(x_d)) 
\end{equation}
In other words, a copula allows us to describe the joint distribution of $(X_1,\cdots,X_d)$ using only their marginals. This gives a lot of flexibility when modelling high dimensional datasets, as we can model each dimension separately, and there is a guaranteed way to link the marginal distributions together to form the joint distribution.

Sklar also shows that if $F$ has univariate marginal distributions $F_1,..., F_d$, then there exists a copula $C(.)$ such that equation \ref{sklar1} holds. Furthermore, if the marginals are continuous, then $C(.)$ can be uniquely determined.

Similarly, by substituting the inverse of equation \ref{inverse relationship} into equation \ref{cop_main}, we obtain the copula equation expressed in terms of the joint CDF and inverse CDFs.
\begin{align}\label{sklar}
    C(\textbf{u}) & = \mathbb{P}(F_{X_1}(X_1) \leq u_1, \cdots, F_{X_d}(X_d) \leq u_d) \nonumber\\
    & = \mathbb{P}(X_1 \leq F_{X_1}^{-1}(X_1), \cdots, F_{X_d}^{-1}(X_d)) \nonumber\\
    & = F_{\textbf{X}}(F_{X_d}^{-1}(u_1),\cdots,F_{X_d}^{-1}(u_d))
\end{align}

Together, these two results are known as Sklar's Theorem, and they guarantee the exists of a copula for any given multivariate CDF with continuous marginals, and provides a closed form equation for how the copula can be constructed.

\subsection{Empirical Copula}
COPOD uses a nonparametric approach based on fitting empirical cumulative distribution functions (ECDFs) called \textbf{Empirical Copula}. Let $X$ be a $d$-dimensional dataset with $n$ observations. We use $X_{j,i}$ to denote the $i^{th}$ observation of the $j^{th}$ dimension. When there is no confusion, we will only use one of the two subscripts, $X_j$ or $X_i$, to denote dimension or observation, respectively. The empirical CDF, $\hat{F}(x)$ is defined:
\begin{equation}\label{empirical cdf}
    \hat{F}(x) = \mathbb{P}((-\infty, x]) = \frac{1}{n}\sum_{i=1}^n\mathbb{I}(X_i \leq x)
\end{equation}

By leveraging the inverse of equation \ref{inverse relationship}, we can obtain the empirical copula observations, $\hat{U}^i$, by
\begin{equation}\label{empirical cop obs}
    (\hat{U}_{1,i},...,\hat{U}_{d,i}) = 
    \big(\hat{F}_1(X_{1,i}),...,\hat{F}_d(X_{d,i})\big)
\end{equation}

Finally, by substituting the empirical copula observations into the first equality of equation \ref{sklar}, we have
\begin{equation}\label{empirical cop}
    \hat{C}(u_1,\cdots,u_d) = \frac{1}{n} \sum_{i=1}^n \mathbb{I}\left(\tilde{U}_{1,i}\leq u_1,\cdots,\tilde{U}_{d,i}\leq u_d\right)
\end{equation}

Nelsen \cite{nelsen2007introduction} show that an empirical copula, $\hat{C}(\textbf{u)}$ with multivariate CDFs supported on $n$ points in the grid $\{1/n, 2/n,\dots, 1\}^d$, has discrete uniform marginals on $\{1/n, 2/n,\dots,1\}$, and asymptotically converges to $C(\textbf{u})$ as a result of the central limit theorem.

\subsection{Theoretical Framework of COPOD}
Outlier detection with COPOD is a three-stage process. First, we compute the empirical CDFs based on the dataset of interest $X = [X_{1,i},\cdots,X_{d,i}], \, i=1,\cdots,n$. Secondly, we use the empirical CDFs to produce the empirical copula function. Finally, we use the empirical copula to approximate the tail probability, namely equation \ref{sklar1} or alternatively, $F_X(x_i) = \mathbb{P}(X_1 \leq x_{1,i},...,X_d \leq x_{d,i})$ for each $x_i$.

\subsubsection{Outliers as Tail Events}
For every observation $x_i$, our goal is to compute the probability of observing a point at least as extreme as $x_i$. In other words, assume that $x_i$ is distributed according to some $d$-variate distribution function $F_X$, we want to calculate $F_X(x_i) = \mathbb{P}(X \leq x_i)$ and $1-F_X(x_i) = \mathbb{P}(X \geq x_i)$. If $x_i$ is an outlier, it should occur infrequently, and as such, the probability of observing a point at least as extreme as $x_i$ should be small. Therefore, if either $F_X(x_i)$ or $1 - F_X(x_i)$ is extremely small, there is evidence that this point has rare occurrence, and is, therefore, likely to be an outlier. We call $F_X(x_i)$ the \textit{left tail probability} of $x_i$ and $1 - F_X(x_i)$ the \textit{right tail probability} of $x_i$. We say that an outlier has a small tail probability if either of the two quantities are small.

In order to compute the tail probabilities empirically, we first compute the empirical CDF using equation \ref{empirical cdf}, once we have the empirical CDFs, we can approximate the inverse of equation \ref{inverse relationship} to obtain the estimated copula observations $u_j$ by feeding each $x_j$ into $\hat{F}_j$. The estimated copula observations tells us the probability of observing something as extreme as $x$ along the $j^{th}$ dimension. Finally, multiplying all $u_j$'s together gives the empirical estimate of the left tail probability.

\subsubsection{Computing Right Tail Probabilities}

In a similar fashion, we can compute the right tail probabilities. All we needed is
\begin{equation}\label{cop inverse}
    C(1-u) = \mathbb{P}(U_1 \geq u_1,...,U_d \geq u_d)
\end{equation}

and since we are feeding $1 - u_j$ into $\hat{F}_j$, this should give us the desired result.

However, because $\hat{F}_j$ is not a true cumulative distribution function, and it only has support from $\{1/n, 2/n,..., 1\}^d$. The largest $u_j$, in this case, would equal to 1, and substituting that into equation \ref{cop inverse} implies that $\mathbb{P}(U_1 \geq u_1,..., U_j \geq u_j,...,U_d \geq u_d) = \mathbb{P}(U_1 \geq u_1,..., U_j \geq 1,...,U_d \geq u_d) = 0$, as no uniforms can take on values greater than or equal to zero. A zero tail probability would make computing the outlier score difficult in subsequent steps.

A simply way to address this challenge is to calculate $\hat{F}_j$ based on $-X = [-X_{1,i},-X_{2,i},\dots,-X_{d,i}]$, and by calculating $\hat{F}_d(-x)$. The desired result follows from substituting $-X_i$ and $-x$ in the last part of equation \ref{empirical cdf}. To avoid confusion in notations, we use $\hat{F}_j$ to denote the left tail ECDF derived from $X$ and $\hat{\bar{F}}_j$ to denote the right tail ECDF derived from $-X$.

\subsubsection{Diminishing Tail Probabilities}
It is well known that various unusual phenomena arise in high-dimensional spaces, which do not occur in low-dimensional settings. This is known as the \textit{curse of dimensionality} and outlier detection algorithms are known to be victims. Consider equation \ref{empirical cop} in both low dimensional and high dimensional settings. As dimensionality increases, the probability of $\tilde{U}_{j,i} \leq u_j, \forall j$ decreases exponentially. Specifically,
\small
\begin{align}\label{cop2}
    \hat{C}(u_1,\dots,u_d) & = \frac{1}{n} \sum_{i=1}^n \mathbb{I}\left(\tilde{U}_{1,i}\leq u_1,\dots,\tilde{U}_{d,i}\leq u_d\right) \nonumber \\
    & = \frac{1}{n} \sum_{i=1}^n \mathbb{I}(\tilde{U}_{1,i}\leq u_1) \times \dots \times \mathbb{I}(\tilde{U}_{d,i}\leq u_d) \nonumber \\
    & \approx \mathbb{P}(\tilde{U}_{1,i}\leq u_1) \times...\times \mathbb{P}(\tilde{U}_{d,i}\leq u_d)
\end{align}
\normalsize
and copula function approaches zero as $d$ increases. In order to avoid diminishing tail probabilities, and using the monotonicity property of the $\log()$ function, we instead work with the sum of the negative log probabilities.
\small
\begin{align} \label{neg log prob}
    -\log(\hat{C}(\textbf{u})) & = 
    -\log(\mathbb{P}(\tilde{U}_{1,i}\leq u_1) \times...\times \mathbb{P}(\tilde{U}_{d,i}\leq u_d)) \nonumber \\
    & = - \sum_{j=1}^d \log(\mathbb{P}(\tilde{U}_{j,i}\leq u_j)) = - \sum_{j=1}^d \log(u_j)
\end{align}
\normalsize
The last line holds since $\tilde{U}_{1,i}$ is uniformly distributed on $[0,1]$.Notice that the motivation of taking logs is similar to that of the log likelihood functions.

\subsubsection{Skewness Correction}
We point out a need for correcting for skewness, in cases where outliers fall on one extreme end of the distribution. Consider Figure 1, where five sets of training and testing results are shown. The images on the left most column are plots of two generated 2-dimensional datasets, with outliers labelled as yellow in the training set and red in the test set. The datasets are generated by taking samples from a combination of normal and uniform distributions on the square. In the next three columns, we show the results of COPOD if left tail probabilities, right tail probabilities, and averages of both tail probabilities are used, respectively.

Because of the nature of this dataset (i.e., all outliers fall on the left end of the 2-dimensional distribution), using the left tail probabilities works surprising well, because all outliers, by the construction of the dataset, are ``smaller" than normalcy. On the other hand, using the right tail worked extremely poorly, because there are no extremely large outliers, and thus COPOD would have wrongfully captured the relatively large points as outliers. Taking an average of the two tail probabilities (column 4 of Figure 1) compromises the situation, and identifies half of the left tail outliers and misidentifies half of the right tail. We should point out that if the dataset is flipped, and all outliers appear on the top right corner while the inliers appear on the bottom left, then COPOD with right tail ECDFs would have worked perfectly, while left tail ECDFs would perform poorly. Averaging, on the other hand, works if both large and small outliers occur.

To overcome this, we note that the skewness of the dataset plays a major role in whether left tail ECDFs or right tail ECDFs should be used. In this example, both marginals of $X_1$ and $X_2$ skew negatively (i.e., the left tail is longer and the mass of the distribution is concentrated on the right). In this case, it makes sense to use the left tail ECDF. For dimensions that skew positively, using the right tail ECDFs would become a better choice. Referring back to equation \ref{neg log prob}, whether we use $\leq$ or $\geq$ depends on whether dimension $j$ skews negatively or positively. The right most column of Figure 1 demonstrates the outcome of COPOD of tail selection is based on skewness of each dimension, and we can see that it is able to capture the correct outliers. We called this the skewness corrected version of COPOD and in Section \ref{sec:empirical_evaluation}, we demonstrate the performance of different variants of COPOD through simulation studies.

\begin{figure*}[htp]\label{toy_example_pic}
\centering
  \includegraphics[scale=0.425]{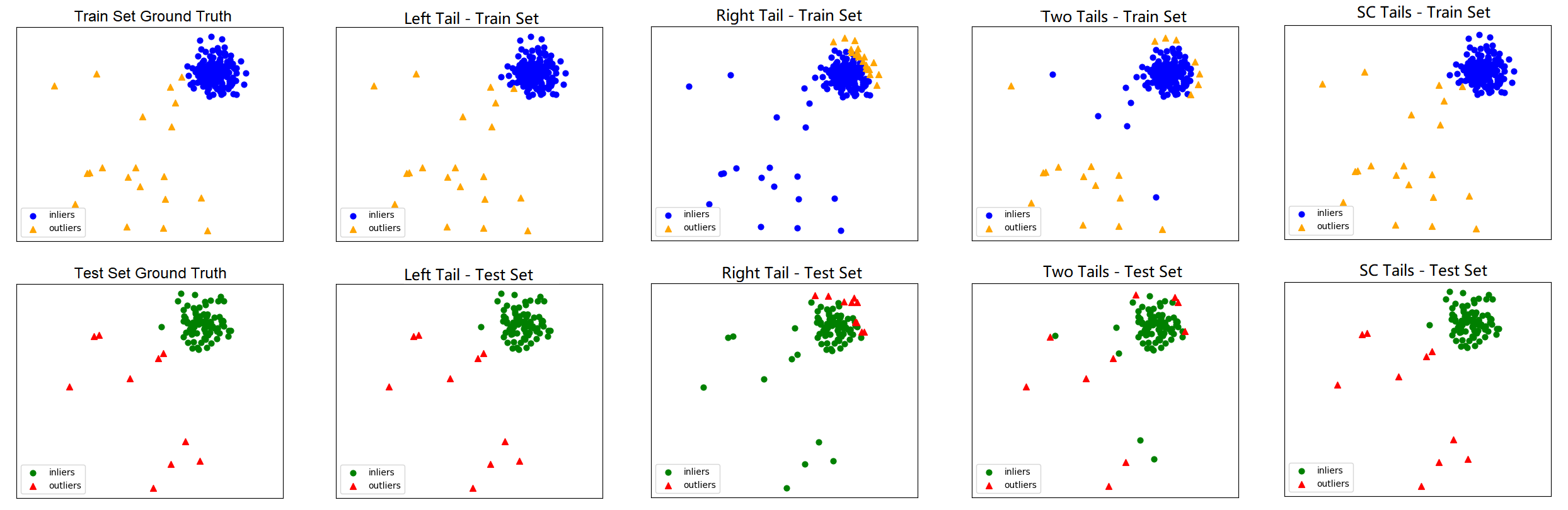}
  \vspace{-0.3in}
  \caption{A toy example of how using different tail probabilities affect the results. The leftmost column are plots of the ground truth, the second column are COPOD's detection if left tail probabilities are used. The middle column corresponds to outlier detection if right tail probabilities are used, followed by both tail probabilities (Two Tails) in the fourth column and skewness corrected tail probabilities (SC Tails) in the rightmost column (which shows the best result).}
\end{figure*}

\subsection{COPOD Algorithm}
Now we are ready to formally introduce COPOD. COPOD is a three stage algorithm, which takes a $d$-dimensional input dataset $X = (X_{1,i},X_{2,i},\dots,X_{d,i}), \, i=1,\dots,n$, and produces an outlier score vector $O(X) = [X_1,...,X_n]$. Outlier scores are between $(0, \infty)$, and are to be use comparatively. \textbf{In other words, the score does not indicate the probability of $X_i$ being an outlier, but rather the relative measure of how likely $X_i$ is when compared to other points in the dataset}. . The bigger $O(X_i)$ is, the more likely $X_i$ is an outlier.

In the first step, COPOD fits $d$ empirical lef tail CDFs, $\hat{F}_1(x),\cdots, \hat{F}_d(x)$, using equation \ref{empirical cdf}, and $d$ empirical right tail CDFs $\hat{\bar{F}}_1(x),\cdots, \hat{\bar{F}}_d(x)$ by replacing $X$ with $-X$. We also compute the skewness vector, $b = [b_1,...,b_d]$, where 
\begin{equation}\label{skewness coef}
        b_i = \frac{\tfrac{1}{n} \sum_{i=1}^n (x_i-\overline{x_i})^3}{ \sqrt{\tfrac{1}{n-1} \sum_{i=1}^n (x_i-\overline{x_i})^2}^{\,3}}
\end{equation}
is the standard formula for estimating skewness.

In the second step, we compute the empirical copula observations for each $X_i$ according to equation \ref{empirical cop obs}, where we obtain $\hat{U}_{d,i} = \hat{F}_d(x_i)$ and $\hat{V}_{d,i} = \hat{\bar{F}}_d(x_i)$, respectively. Further, we calculate the skewness corrected empirical copula observations, $\hat{W}_{d,i} = \hat{U}_{d,i}$ if $b_d < 0$, otherwise $\hat{V}_{d,i}$.

Lastly, we compute the probability of observing a point at least as extreme as each $x_i$ along each dimension. We take the maximum of the negative log of the probability generated by the left tail empirical copula, right tail empirical copula and skewness corrected empirical copula to be the outlier score.

Intuitively, the smaller the tail probability is, the bigger its minus log, and so we claim a point is an outlier if it has a small left tail probability, a small right tail probability, or a small skewness corrected tail probability. We summarise the algorithm in Algorithm \ref{alg:COPOD}.

\begin{algorithm} [ht]
	\caption{Copula Outlier Detector}
	\label{alg:COPOD}
	\scriptsize
	\begin{algorithmic}[1]
		\Require{input data $X$}
		\Ensure{outlier scores $O(X)$}
		\For{each dimension $d$}
		\State Compute left tail ECDFs: $\hat{F}_d(x) = \frac{1}{n}\sum_1^n\mathbb{I}(X_i \leq x)$
		\State Compute right tail ECDFs: $\hat{\bar{F}}_d(x) = \frac{1}{n}\sum_1^n\mathbb{I}(-X_i \leq -x)$
		\State Compute the skewness coefficient according to Equation \ref{skewness coef}.
		\EndFor
		
		\For{each $i$ in $1,...n$}
		\State Compute empirical copula observations
		\Indent
		\State   $\hat{U}_{d,i} = \hat{F}_d(x_i)$, 
		\State   $\hat{V}_{d,i} = \hat{\bar{F}}_d(x_i)$
		\State   $\hat{W}_{d,i} = \hat{U}_{d,i} \: \text{if $b_d < 0$ otherwise} \: \hat{V}_{d,i}$ 
		\EndIndent
	    \State Calculate tail probabilities of $X_i$ as follows:
	    \Indent
    	    \State 	    $p_l = -\sum_{j=1}^d \log(\hat{U}_{j,i})$
    		\State      $p_r = -\sum_{j=1}^d \log(\hat{V}_{j,i})$
    	    \State 	    $p_s = -\sum_{j=1}^d \log(\hat{W}_{j,i})$
		\EndIndent
	    \State Outlier Score $O(x_i) = \max\{p_l, \; p_r, \; p_s\}$
		\EndFor
		\State \textbf{Return} $O(X)=[O(x_1),...,O(x_d)]^T$
	\end{algorithmic}
\end{algorithm}

\subsection{COPOD as an Interpretable Outlier Detector}
Interpretability in machine learning is an important concept, as it provides domain experts some insights into how algorithms make their decisions. Interpretable algorithms provide both transparency and reliability \cite{hu2019optimal}. Having a transparent model means that humans can learn from the thought process of models, and try to discover the ``whys" behind why a particular data point is classified. Reliability, on the other hand, ensures that machine learning models can be properly audited so that their reasoning processes align with human expectations.

As COPOD evaluates the anomalous behaviours on a dimensional basis, it is possible to use COPOD as an explainable detector. From equation \ref{alg:COPOD}, we know that if a point has a large outlier score, then at least one of the three tail copulas is large. 

Thus, let $O_d(x_i) = \max\{\log(\hat{U}_{d,i}), \: \log(\hat{V}_{d,i}), \: \log(\hat{W}_{d,i})\}$ be the outlier score for dimension $d$, and using fact that the $\log$ function is monotone, we can see that each $O_d(x_i)$ represents the degree of anomalousness of  dimension $d$. This can be compared against some desired threshold, such as $-\log(0.01) = 4.61$, or top $\alpha$ percent of $O_d(x_i) \; \forall i$ to give practitioners some indications why particular points are considered outliers.

We illustrate the interpretability aspect of COPOD with the Breast Cancer Wisconsin (Diagnostic) Data Set (BreastW) as an example. Cell samples are provided by Dr. William H. Wolberg from the University of Wisconsin as part of his reports on clinical cases. This batch of data used has 369 samples, with 9 features listed below, and two outcome classes (benign and malignant). Features are scored on a scale of 1-10, and are 1) Clump Thickness, 2) Uniformity of Cell Size, 3) Uniformity of Cell Shape, 4) Marginal Adhesion, 5) Single Epithelial Cell Size, 6) Bare Nuclei, 7) Bland Chromatin, 8) Normal Nucleoli and 9) Mitoses. Section \ref{sec:empirical_evaluation} discusses this dataset in greater detail.

We are particularly interested in providing an explanation, in addition to giving the right classification of outlier vs. inliner, as this can provide useful guidance for physicians to further investigate why certain cells are potentially harmful. We illustrate how COPOD attempts to explain its outlier detection process using the following example.

For each of the 9 dimensions of data point $i$, we plot $O_d(x_i)$, the dimensional outlier score. In additional, we also plot the $99^{th}$ percentile band in green, and the $65.01^{th}$ percentile band in black (which corresponds to $1 -$ contamination rate). We call this the \textit{Dimensional Outlier Graph}. In the analysis below, we plot the Dimensional Outlier Graph for data point 70 (row 70 in BreastW), which is an outlier and has been successfully detected by COPOD.

\begin{figure}[tp]\label{od70}
    \includegraphics[width=0.5\textwidth]{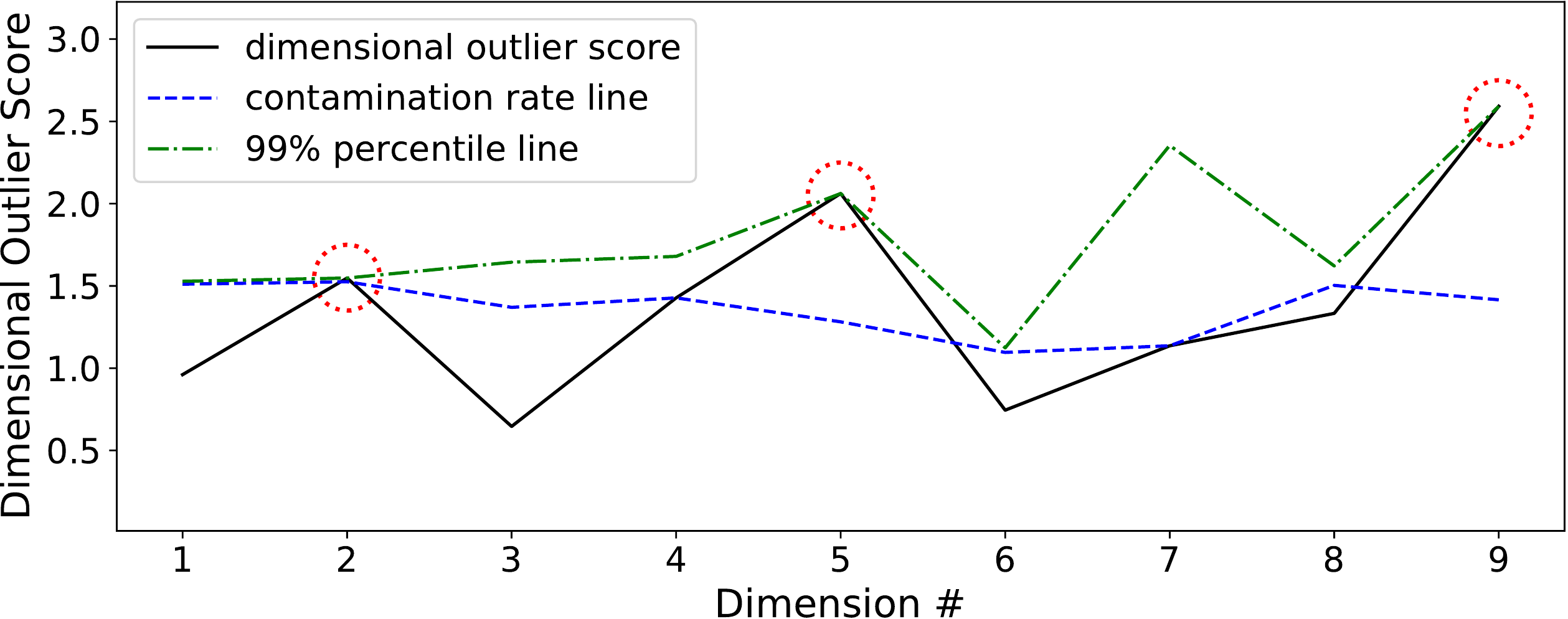}
    \vspace{-0.3in}
    \caption{\textit{Dimensional Outlier Graph} for sample 70 (row 70); dimension 2, 5, 9 are identified as outlier subspaces.}
\end{figure}

We see that for features 2, 5 and 9, the dimensional outlier scores have touched the $99^{th}$ percentile band, while all other features remain below the contamination rate band. This suggests that this point is likely an outlier because it is extremely non-uniform in size, has a large single epithelial cell size, and is more likely to reproduce via mitosis rather than meiosis.

\section{Empirical Evaluation}
\label{sec:empirical_evaluation}
\subsection{Datasets and Evaluation Metrics} 
30 public outlier detection benchmark datasets used in this study are from \texttt{ODDS}\footnote{ODDS Library: \url{http://odds.cs.stonybrook.edu}} and \texttt{DAMI} Datasets\cite{dami}.
In each experiment, 60\% of the data is used for training and the remaining 40\% is set aside for validation. Performance is evaluated by taking the average score of 10 independent trials using area under the receiver operating characteristic (ROC-AUC) and average precision (AP).

\subsection{COPOD Implementation}
Our implementation of COPOD\footnote{\url{https://github.com/winstonll/COPOD}} is featured and included in PyOD \cite{zhao2019pyod}, a popular open-source Python toolbox for performing scalable outlier detection on multivariate data with a single, well-documented API. This allows us to easily compare COPOD with other outlier detectors. In the subsequent experiments, a Windows laptop with Intel i5-8265U @ 1.60 GHz quad core CPU and 8 GB of memory is used.

\subsection{Performance of COPOD variants}
In the first experiment, we compare performances of a several variants of COPOD. Specifically, we look at comparisons between four variants, namely, \textit{left tail ECDF}, \textit{right tail ECDF}, \textit{two tails ECDF}, and \textit{skewness corrected ECDF} as outlined in Section \ref{sec:proposed_algorithm}. 
COPOD with skewness corrected ECDF achieves the best performance, with an average ROC-AUC of 0.8247 and average precision of 0.5649, followed by COPOD with two tails (ROC: 0.7818, AP: 0.4767), COPOD with right tail ECDF (ROC: 0.7148, AP: 0.4118) and finally COPOD with left tail ECDF (ROC: 0.5061, AP: 0.2514).

\subsection{Performance vs. other outlier detectors}
\begin{table*}[ht]
\setlength{\tabcolsep}{3pt}
\centering
	\caption{ROC-AUC scores of detector performance (average of 10 independent trials, highest score highlighted in bold); rank is shown in parenthesis (lower is better). COPOD outperforms all baselines.}
\vspace{-0.1in}
\footnotesize
\scalebox{0.735}{
\begin{tabular}{l|rrrrrrrrr|r}
\toprule
\textbf{Data} & \textbf{ABOD} & \textbf{CBLOF} & \textbf{FB} & \textbf{HBOS} & \textbf{IForest} & \textbf{KNN} & \textbf{LODA} & \textbf{LOF} & \textbf{OCSVM} & \textbf{COPOD} \\
\midrule
Arrhythmia (mat) & 0.7688 (9) & 0.7838 (6) & 0.7839 (5) & \textbf{0.8219 (1)} & 0.802 (3) & 0.7861 (4) & 0.7497 (10) & 0.7787 (8) & 0.7812 (7) & 0.8021 (2) \\
Breastw (mat) & 0.3959 (10) & 0.9651 (6) & 0.396 (9) & 0.9825 (4) & 0.9869 (3) & 0.9755 (5) & 0.9873 (2) & 0.47 (8) & 0.9612 (7) & \textbf{0.9936 (1)} \\
Cardio (mat) & 0.5692 (10) & 0.81 (6) & 0.7057 (8) & 0.8351 (5) & 0.9226 (2) & 0.7236 (7) & 0.8559 (4) & 0.5736 (9) & \textbf{0.9348 (1)} & 0.8974 (3) \\
Ionosphere (mat) & 0.9248 (3) & 0.8972 (4) & \textbf{0.9495 (1)} & 0.5614 (10) & 0.8472 (6) & 0.9267 (2) & 0.7975 (9) & 0.8753 (5) & 0.8419 (7) & 0.8307 (8) \\
Lympho (mat) & 0.911 (8) & 0.9673 (7) & 0.8809 (9) & \textbf{0.9957 (1)} & 0.9934 (3) & 0.9745 (6) & 0.7303 (10) & 0.9771 (4) & 0.9759 (5) & 0.9935 (2) \\
Mammography (mat) & 0.5494 (10) & 0.8213 (8) & \textbf{0.8775 (2)} & 0.8342 (7) & 0.8602 (5) & 0.841 (6) & 0.8809 (3) & 0.7293 (9) & 0.8724 (4) & \textbf{0.8942 (1)} \\
Optdigits (mat) & 0.4667 (7) & 0.7692 (3) & 0.7948 (2) & \textbf{0.8732 (1)} & 0.7209 (5) & 0.3708 (10) & 0.3982 (9) & 0.45 (8) & 0.4997 (6) & 0.7328 (4) \\
Pima (mat) & 0.6794 (3) & 0.6578 (6) & 0.3781 (10) & 0.7000 (2) & 0.6778 (4) & \textbf{0.7078 (1)} & 0.592 (9) & 0.6271 (7) & 0.6215 (8) & 0.6638 (5) \\
Satellite (mat) & 0.5714 (9) & 0.7494 (2) & 0.6525 (7) & \textbf{0.7581 (1)} & 0.7024 (3) & 0.6836 (4) & 0.5969 (8) & 0.5573 (10) & 0.6622 (5) & 0.6612 (6) \\
Satimage-2 (mat) & 0.819 (8) & \textbf{0.9992 (1)} & 0.5670 (9) & 0.9804 (6) & 0.9947 (3) & 0.9536 (7) & 0.9866 (4) & 0.4577 (10) & 0.9978 (2) & 0.9852 (5) \\
Shuttle (mat) & 0.6591 (7) & 0.6075 (8) & 0.6715 (6) & 0.9948 (3) & \textbf{0.9983 (1)} & 0.7381 (5) & 0.5892 (9) & 0.5243 (10) & 0.9945 (4) & 0.998 (2) \\
Speech (mat) & \textbf{0.6267 (1)} & 0.4515 (7) & 0.4503 (8) & 0.4538 (6) & 0.447 (10) & 0.4581 (4) & 0.4551 (5) & 0.4714 (3) & 0.4472 (9) & 0.4845 (2) \\
WBC (mat) & 0.9047 (9) & 0.9201 (7) & 0.4615 (10) & 0.9516 (2) & 0.9314 (6) & 0.9366 (3) & 0.9126 (8) & 0.9349 (4) & 0.9319 (5) & \textbf{0.9747 (1)} \\
Wine (mat) & 0.4305 (9) & 0.2835 (10) & 0.9346 (2) & 0.9001 (4) & 0.8163 (5) & 0.5177 (8) & 0.7873 (6) & 0.9045 (3) & 0.6363 (7) & \textbf{0.949 (1)} \\
Arrhythmia (arff) & 0.7396 (9) & 0.7514 (5) & 0.7614 (2) & 0.7493 (7) & 0.7567 (3) & 0.7508 (6) & 0.7028 (10) & 0.749 (8) & 0.7557 (4) & \textbf{0.7618 (1)} \\
Cardiotocography (arff) & 0.4593 (10) & 0.5673 (6) & 0.5184 (7) & 0.5936 (5) & 0.6828 (2) & 0.4904 (9) & 0.6738 (3) & 0.5045 (8) & \textbf{0.6976 (1)} & 0.637 (4) \\
HeartDisease (arff) & 0.5969 (6) & 0.5926 (7) & 0.6286 (3) & \textbf{0.6893 (1)} & 0.602 (5) & 0.6136 (4) & 0.4821 (10) & 0.5626 (8) & 0.5548 (9) & 0.6728 (2) \\
Hepatitis (arff) & 0.7155 (9) & 0.7585 (8) & 0.8420 (2) & 0.7961 (5) & 0.7935 (6) & 0.8111 (4) & 0.6075 (10) & 0.8305 (3) & 0.7644 (7) & \textbf{0.8432 (1)} \\
InternetAds (arff) & 0.6427 (4) & 0.6166 (7) & 0.6132 (8) & \textbf{0.6994 (1)} & 0.6851 (2) & 0.6216 (6) & 0.5276 (10) & 0.6057 (9) & 0.6233 (5) & 0.6763 (3) \\
Ionosphere (arff) & \textbf{0.9250 (1)} & 0.8859 (3) & 0.8800 (4) & 0.552 (10) & 0.836 (7) & 0.9187 (2) & 0.8222 (8) & 0.8604 (5) & 0.8381 (6) & 0.8219 (9) \\
KDDCup99 (arff) & 0.6925 (8) & 0.9970 (2) & 0.391 (10) & 0.995 (4) & 0.9968 (3) & 0.7611 (6) & 0.7277 (7) & 0.5774 (9) & 0.9948 (5) & \textbf{0.9972 (1)} \\
Lymphography (arff) & 0.9864 (9) & 0.9956 (6) & 0.9961 (5) & 0.9984 (3) & 0.9994 (1) & 0.9965 (4) & 0.8385 (10) & 0.995 (7) & 0.9915 (8) & 0.9985 (2) \\
Pima (arff) & 0.6666 (3) & 0.6658 (4) & 0.6544 (7) & 0.6591 (5) & 0.6756 (2) & \textbf{0.7063 (1)} & 0.5974 (10) & 0.6582 (6) & 0.6403 (9) & 0.6466 (8) \\
Shuttle (arff) & 0.8338 (7) & 0.9802 (2) & 0.3671 (10) & 0.7924 (8) & 0.8661 (6) & 0.9607 (4) & 0.7066 (9) & \textbf{0.9857 (1)} & 0.9658 (3) & 0.8795 (5) \\
SpamBase (arff) & 0.4349 (9) & 0.5476 (5) & 0.6795 (2) & 0.6698 (3) & 0.6328 (4) & 0.5399 (6) & 0.4449 (8) & 0.4255 (10) & 0.5386 (7) & \textbf{0.7007 (1)} \\
Stamps (arff) & 0.7617 (7) & 0.7299 (10) & 0.7419 (8) & 0.9182 (2) & 0.9132 (3) & 0.8738 (6) & 0.9014 (4) & 0.739 (9) & 0.88 (5) & \textbf{0.9388 (1)} \\
Waveform (arff) & 0.6522 (9) & 0.7143 (4) & 0.6966 (5) & 0.683 (6) & 0.6813 (7) & 0.7403 (2) & 0.622 (10) & 0.7354 (3) & 0.6649 (8) & \textbf{0.7865 (1)} \\
WBC (arff) & 0.9559 (8) & 0.9658 (7) & 0.9432 (9) & 0.9819 (3) & \textbf{0.9906 (1)} & 0.9715 (6) & 0.9811 (4) & 0.9315 (10) & 0.9751 (5) & 0.9904 (2) \\
WDBC (arff) & 0.9032 (9) & 0.9088 (8) & 0.5388 (10) & 0.9674 (2) & 0.9462 (3) & 0.9272 (6) & 0.9455 (4) & 0.9204 (7) & 0.938 (5) & \textbf{0.9826 (1)} \\
WPBC (arff) & 0.457 (10) & 0.4918 (9) & 0.5196 (4) & 0.5357 (2) & 0.5164 (5) & 0.526 (3) & 0.5029 (7) & 0.5108 (6) & 0.4979 (8) & \textbf{0.5465 (1)} \\
\midrule
AVG & 0.6900 (9) & 0.7617 (5) & 0.6792 (10) & 0.7974 (3) & 0.8092 (2) & 0.7601 (6) & 0.7135 (7) & 0.6974 (8) & 0.7826 (4) & \textbf{0.8247 (1)} \\
\bottomrule
\end{tabular}}
	\label{table:od comparison roc} 
\end{table*}

In this section, we compare the performance of COPOD with 9 other outlier detectors. We aim to include a variety of detectors to make the comparison robust. Specifically, the 9 competitors are Angle-Based Outlier Detection (ABOD), Clustering-Based Local Outlier Factor (CBLOF), Feature Bagging (FB), Histogram-based Outlier Score (HBOS), Isolation Forest (IForest), k Nearest Neighbors (KNN), Lightweight On-line Detector of Anomalies (LODA), Local Outlier Factor(LOF) and One-Class Support Vector Machines (OCSVM). Their implementations can all be found in PyOD.

Of the 30 datasets used, COPOD scores the highest in terms of ROC-AUC. Specifically, COPOD achieves an average ROC-AUC score of 82.47\%, which is 1.5\% higher than the second best alternative. COPOD ranks first in 12 out of 30 occasions and ranks in the top two places in 16 out of 30 occasions. 

Similar results are seen for average precision scores. Of the 10 detectors, COPOD achieves an average precision of 56.49\%, which is 2.7\% higher than the second place. COPOD ranks first in 12 out of 30 datasets, and ranks in the top two places in 17 times.

Finally, COPOD ranks third in computation time with an average of 0.227 seconds to process the 30 test dataset, falling just behind HBOS (0.046s) and LODA (0.062s).

\begin{table*}[ht]
\setlength{\tabcolsep}{3pt}
\centering
	\caption{Average precision of detector performance (average of 10 independent trials, highest score highlighted in bold); rank is shown in parenthesis (lower is better). COPOD outperforms all baselines.}
\vspace{-0.1in}
\footnotesize
\scalebox{0.735}{
\begin{tabular}{l|rrrrrrrrr|r}
\toprule
\textbf{Data} & \textbf{ABOD} & \textbf{CBLOF} & \textbf{FB} & \textbf{HBOS} & \textbf{IForest} & \textbf{KNN} & \textbf{LODA} & \textbf{LOF} & \textbf{OCSVM} & \textbf{COPOD} \\
\midrule
Arrhythmia (mat) & 0.3585 (10) & 0.399 (6) & 0.3872 (8) & 0.4928 (2) & \textbf{0.5062 (1)} & 0.3968 (7) & 0.4359 (4) & 0.3742 (9) & 0.4045 (5) & 0.4727 (3) \\
Breastw (mat) & 0.2945 (9) & 0.9141 (7) & 0.2927 (10) & 0.953 (4) & 0.9721 (3) & 0.9269 (6) & 0.9777 (2) & 0.3215 (8) & 0.9343 (5) & \textbf{0.9877 (1)} \\
Cardio (mat) & 0.1936 (9) & 0.4136 (6) & 0.2091 (8) & 0.4602 (4) & 0.5763 (2) & 0.3451 (7) & 0.4237 (5) & 0.1626 (10) & 0.5324 (3) & \textbf{0.5793 (1)} \\
Ionosphere (mat) & 0.9141 (2) & 0.8712 (3) & 0.1198 (10) & 0.3664 (9) & 0.7889 (6) & \textbf{0.9238 (1)} & 0.7156 (8) & 0.8211 (5) & 0.8229 (4) & 0.7194 (7) \\
Lympho (mat) & 0.5168 (9) & 0.8078 (8) & 0.8266 (4) & 0.9252 (2) & \textbf{0.9516 (1)} & 0.82 (6) & 0.4165 (10) & 0.8252 (5) & 0.8144 (7) & 0.8936 (3) \\
Mammography (mat) & 0.0232 (10) & 0.1369 (7) & 0.4258 (2) & 0.1222 (8) & 0.2329 (4) & 0.1695 (6) & 0.2808 (3) & 0.1179 (9) & 0.188 (5) & \textbf{0.429 (1)} \\
Optdigits (mat) & 0.0279 (7) & 0.0624 (3) & 0.1092 (2) & \textbf{0.1957 (1)} & 0.0553 (4) & 0.0223 (10) & 0.0244 (9) & 0.0288 (6) & 0.0275 (8) & 0.0532 (5) \\
Pima (mat) & 0.5107 (4) & 0.4771 (6) & 0.0234 (10) & \textbf{0.5693 (1)} & 0.5031 (5) & \textbf{0.5145 (3)} & 0.4081 (9) & 0.4299 (8) & 0.4608 (7) & 0.5407 (2) \\
Satellite (mat) & 0.3973 (9) & \textbf{0.6897 (1)} & 0.4518 (8) & 0.6866 (2) & 0.6539 (3) & 0.5431 (7) & 0.5805 (6) & 0.3898 (10) & 0.6529 (4) & 0.5854 (5) \\
Satimage-2 (mat) & 0.1874 (9) & \textbf{0.9777 (1)} & 0.3984 (8) & 0.758 (6) & 0.9285 (3) & 0.4186 (7) & 0.8695 (4) & 0.0267 (10) & 0.9746 (2) & 0.8599 (5) \\
Shuttle (mat) & 0.1705 (8) & 0.1945 (7) & 0.0353 (10) & 0.9796 (3) & \textbf{0.9857 (1)} & 0.2036 (6) & 0.3786 (5) & 0.1421 (9) & 0.9015 (4) & 0.9807 (2) \\
Speech (mat) & \textbf{0.0395 (1)} & 0.0216 (5) & 0.0223 (4) & 0.0272 (2) & 0.0177 (9) & 0.0216 (5) & 0.0165 (10) & 0.0243 (3) & 0.0214 (7) & 0.0195 (8) \\
WBC (mat) & 0.3545 (10) & 0.4995 (9) & 0.562 (5) & 0.6627 (2) & 0.5898 (3) & 0.5294 (7) & 0.5641 (4) & 0.558 (6) & 0.5137 (8) & \textbf{0.7827 (1)} \\
Wine (mat) & 0.0838 (9) & 0.0598 (10) & 0.3563 (4) & 0.405 (2) & 0.2794 (5) & 0.0946 (8) & 0.2783 (6) & 0.3611 (3) & 0.1414 (7) & \textbf{0.6082 (1)} \\
Arrhythmia (arff) & 0.6987 (9) & 0.7118 (6) & 0.7314 (4) & 0.7504 (2) & 0.7463 (3) & 0.7118 (6) & 0.697 (10) & 0.704 (8) & 0.7164 (5) & \textbf{0.7519 (1)} \\
Cardiotocography (arff) & 0.247 (10) & 0.3625 (6) & 0.2702 (8) & 0.3664 (5) & \textbf{0.4342 (1)} & 0.3108 (7) & 0.4315 (2) & 0.2582 (9) & 0.4185 (3) & 0.3778 (4) \\
HeartDisease (arff) & 0.5341 (5) & 0.521 (7) & 0.5506 (3) & 0.6247 (2) & 0.5337 (6) & 0.5384 (4) & 0.445 (10) & 0.4779 (9) & 0.5128 (8) & \textbf{0.64 (1)} \\
Hepatitis (arff) & 0.3304 (10) & 0.3739 (9) & \textbf{0.5660 (2)} & 0.4732 (5) & 0.4424 (6) & 0.4752 (4) & 0.3957 (8) & 0.4986 (3) & 0.4274 (7) & \textbf{0.5849 (1)} \\
InternetAds (arff) & 0.276 (7) & 0.3153 (5) & 0.2737 (8) & \textbf{0.5347 (1)} & 0.4895 (3) & 0.2805 (6) & 0.2417 (10) & 0.2622 (9) & 0.3157 (4) & 0.5102 (2) \\
Ionosphere (arff) & \textbf{0.5055 (6)} & 0.4839 (7) & 0.5286 (2) & 0.5281 (3) & 0.5146 (5) & 0.525 (4) & 0.4199 (10) & 0.4584 (9) & 0.4779 (8) & \textbf{0.5302 (1)} \\
KDDCup99 (arff) & 0.3798 (7) & 0.4143 (5) & 0.3018 (10) & 0.5249 (2) & 0.4923 (3) & 0.4219 (4) & 0.3752 (8) & 0.3489 (9) & 0.4115 (6) & \textbf{0.5667 (1)} \\
Lymphography (arff) & 0.2465 (8) & 0.2409 (9) & 0.4352 (2) & 0.3983 (4) & 0.3936 (5) & 0.3412 (7) & 0.4038 (3) & 0.2291 (10) & 0.3464 (6) & \textbf{0.4637 (1)} \\
Pima (arff) & \textbf{0.9153 (1)} & 0.8619 (3) & 0.4545 (9) & 0.3637 (10) & 0.7767 (6) & \textbf{0.9145 (2)} & 0.7631 (7) & 0.811 (5) & 0.8215 (4) & 0.7031 (8) \\
Shuttle (arff) & 0.0182 (10) & 0.1977 (5) & \textbf{0.3171 (1)} & 0.2781 (2) & 0.2733 (3) & 0.0457 (8) & 0.1352 (6) & 0.028 (9) & 0.1249 (7) & 0.2389 (4) \\
SpamBase (arff) & 0.8013 (8) & 0.925 (6) & 0.4128 (10) & 0.9783 (3) & \textbf{0.9917 (1)} & 0.9417 (4) & 0.4914 (9) & 0.9254 (5) & 0.8367 (7) & 0.9817 (2) \\
Stamps (arff) & 0.2626 (5) & 0.3583 (3) & 0.1195 (8) & 0.0804 (10) & 0.1297 (7) & 0.3597 (2) & 0.1316 (6) & \textbf{0.3945 (1)} & 0.3042 (4) & 0.0854 (9) \\
Waveform (arff) & 0.0656 (7) & 0.1293 (3) & \textbf{0.1925 (1)} & 0.0547 (9) & 0.0559 (8) & 0.1343 (2) & 0.0521 (10) & 0.108 (4) & 0.0702 (6) & 0.0785 (5) \\
WBC (arff) & 0.5422 (9) & 0.5561 (8) & 0.7534 (3) & 0.6935 (6) & \textbf{0.863 (1)} & 0.5806 (7) & 0.7242 (4) & 0.3303 (10) & 0.7018 (5) & 0.8381 (2) \\
WDBC (arff) & 0.43 (9) & 0.6669 (6) & 0.2973 (10) & 0.7952 (2) & 0.7178 (4) & 0.6535 (7) & 0.7935 (3) & 0.7038 (5) & 0.6123 (8) & \textbf{0.8407 (1)} \\
WPBC (arff) & 0.2095 (10) & 0.2244 (7) & \textbf{0.2412 (2)} & 0.2303 (5) & 0.2313 (4) & 0.2328 (3) & 0.2231 (8) & 0.2248 (6) & 0.2226 (9) & \textbf{0.2438 (1)} \\
\midrule
AVG & 0.3512 (10) & 0.4623 (5) & 0.3715 (9) & 0.5093 (3) & 0.5376 (2) & 0.4466 (6) & 0.4365 (7) & 0.3782 (8) & 0.4904 (4) & \textbf{0.5649 (1)}\\
\bottomrule
	\end{tabular}}
	\label{table:od comparison map} 
\end{table*}

\subsection{Scaling to High Dimensions}
COPOD is an efficient algorithm that scales well in high dimensional settings. Unlike proximity based models that require pairwise distance calculation \cite{Zhao2018,zhao2020suod}, or learning based models that require training, COPOD incurs low computational overhead. We carry out an experiment using random datasets of dimensions 10, 100, 1,000, and 10,000. The number of observations are 1,000, 10,000, 100,000, and 1,000,000. All datasets are randomly generated and used strictly for evaluating COPOD's computation time as shown in Table \ref{table:copod speed}.

\begin{table}[H]
\caption{COPOD's Performance in Seconds Under Different Dimension and Data Size. It scales well regarding dimensions.}
\vspace{-0.1in}
\centering
\scalebox{0.9}{
\begin{tabular}{l|llll}
\toprule
 & d=10 & d=100 & d=1,000 & d=10,000 \\
\midrule
n=1,000 & 0.0687 & 0.1731 & 1.1632 & 11.4605 \\
n=10,000 & 0.1718 & 0.4686 & 5.2441 & 55.1901 \\
n=100,000 & 0.6405 & 7.1858 & 70.5410 & 567.1058 \\
n=1,000,000 & 11.4036 & 130.9741 & 694.4056 & 5376.5937\\
\bottomrule
\end{tabular}}
\label{table:copod speed}
\end{table}

Table \ref{table:copod speed} outlines COPOD's performance and its scalability across both number of dimensions and number of data points. Even on a moderately equipped personal computer, COPOD can easily handle datasets with 10,000 dimensions and 1,000,000 data points in just under an hour and half. In addition, COPOD requires no re-training fit new data points, and thus is desirable for real time prediction applications.

\section{Conclusion}
\label{sec:conclusion}
In this paper, we present COPOD, a novel outlier detection algorithm based on estimating tail probabilities using empirical copula. To the best of our knowledge, this is the first time that copula models have been used in explainable outlier detection tasks. Our key contributions are 1) propose a novel, parameter-free, copula-based outlier detection algorithm, 2) perform comparative studies on 30 datasets to show that COPOD outperforms in most cases and is also one of the fastest algorithms, and 3) release an easy-to-use Python implementation for accessibility and reproducibility. Through extensive experiments with real world data, we show that COPOD's performance is comparable or better to leading outlier detection algorithms, regarding both detection accuracy and computational cost.

\bibliographystyle{./bibliography/IEEEtran}
\bibliography{./bibliography/cod}

\begin{thebibliography}{10}
\providecommand{\url}[1]{#1}
\csname url@samestyle\endcsname
\providecommand{\newblock}{\relax}
\providecommand{\bibinfo}[2]{#2}
\providecommand{\BIBentrySTDinterwordspacing}{\spaceskip=0pt\relax}
\providecommand{\BIBentryALTinterwordstretchfactor}{4}
\providecommand{\BIBentryALTinterwordspacing}{\spaceskip=\fontdimen2\font plus
\BIBentryALTinterwordstretchfactor\fontdimen3\font minus
  \fontdimen4\font\relax}
\providecommand{\BIBforeignlanguage}[2]{{%
\expandafter\ifx\csname l@#1\endcsname\relax
\typeout{** WARNING: IEEEtran.bst: No hyphenation pattern has been}%
\typeout{** loaded for the language `#1'. Using the pattern for}%
\typeout{** the default language instead.}%
\else
\language=\csname l@#1\endcsname
\fi
#2}}
\providecommand{\BIBdecl}{\relax}
\BIBdecl

\bibitem{zhao2019pyod}
Y.~Zhao, Z.~Nasrullah, and Z.~Li, ``{PyOD}: A python toolbox for scalable
  outlier detection,'' \emph{Journal of Machine Learning Research}, 2019.

\bibitem{wan2020sync}
C.~Wan, Z.~Li, A.~Guo, and Y.~Zhao, ``Syn{C}: A unified framework for
  generating synthetic population with gaussian copula,'' \emph{Workshops at
  the Thirty-Fourth AAAI Conference on Artificial Intelligence}, 2020.

\bibitem{mcd}
{ P. J. Rousseeuw and K. V. Driessen}, ``A fast algorithm for the minimum
  covariance determinant estimator,'' \emph{Technometrics}, no. 41(3), 1999.

\bibitem{Ramaswamy2000efficient}
S.~Ramaswamy, R.~Rastogi, and K.~Shim, ``Efficient algorithms for mining
  outliers from large data sets,'' in \emph{ACM SIGMOD Record}, 2000.

\bibitem{scholkopf2001estimating}
B.~Sch{\"o}lkopf, J.~C. Platt, J.~Shawe-Taylor, A.~J. Smola, and R.~C.
  Williamson, ``Estimating the support of a high-dimensional distribution,''
  \emph{Neural Computation}, vol.~13, no.~7, pp. 1443--1471, 2001.

\bibitem{zhao2019lscp}
Y.~Zhao, Z.~Nasrullah, M.~K. Hryniewicki, and Z.~Li, ``{LSCP:} locally
  selective combination in parallel outlier ensembles,'' in \emph{SDM}, 2019.

\bibitem{Liu2019generative}
Y.~Liu, Z.~Li, C.~Zhou, Y.~Jiang, J.~Sun, M.~Wang, and X.~He, ``Generative
  adversarial active learning for unsupervised outlier detection,''
  \emph{TKDE}, 2019.

\bibitem{li2020autood}
Y.~Li, Z.~Chen, D.~Zha, K.~Zhou, H.~Jin, H.~Chen, and X.~Hu, ``Autood:
  Automated outlier detection via curiosity-guided search and self-imitation
  learning,'' \emph{arXiv preprint arXiv:2006.11321}, 2020.

\bibitem{sklar1959fonctions}
A.~Sklar, ``Fonctions de repartition an dimensions et leurs marges,''
  \emph{Publ. inst. statist. univ. Paris}, vol.~8, pp. 229--231, 1959.

\bibitem{nelsen2007introduction}
R.~B. Nelsen, \emph{An Introduction to Copulas}.\hskip 1em plus 0.5em minus
  0.4em\relax Springer Science, 2007.

\bibitem{hu2019optimal}
X.~Hu, C.~Rudin, and M.~Seltzer, ``Optimal sparse decision trees,'' in
  \emph{NIPS}, 2019.

\bibitem{dami}
G.~O. Campos, A.~Zimek, J.~Sander, R.~J. G.~B. Campello, B.~Micenková,
  E.~Schubert, I.~Assent, and M.~E. Houle, ``On the evaluation of unsupervised
  outlier detection: measures, datasets, and an empirical study,'' \emph{DMKD},
  vol.~30, no. 9-10, pp. 891--927, 2016.

\bibitem{Zhao2018}
Y.~Zhao and M.~K. Hryniewicki, ``{XGBOD: Improving Supervised Outlier Detection
  with Unsupervised Representation Learning},'' \emph{IJCNN}.

\bibitem{zhao2020suod}
Y.~Zhao, X.~Hu, C.~Cheng, C.~Wang, C.~Xiao, Y.~Wang, J.~Sun, and L.~Akoglu,
  ``Suod: A scalable unsupervised outlier detection framework,'' \emph{arXiv
  preprint arXiv:2003.05731}, 2020.

\end{thebibliography}
\end{document}